\documentclass{article}

\PassOptionsToPackage{numbers}{natbib}
\usepackage[preprint]{neurips_2020}

\usepackage[utf8]{inputenc} 
\usepackage[T1]{fontenc}    
\usepackage{hyperref}       
\usepackage{url}            
\usepackage{booktabs}       
\usepackage{amsfonts}       
\usepackage{nicefrac}       
\usepackage{microtype}      
\usepackage{graphicx}
\usepackage{listings}
\usepackage{color}
\usepackage{parcolumns}
\usepackage{array}
\usepackage{makecell}
\usepackage{multirow}
\usepackage{amssymb}
\usepackage{pifont}
\usepackage{graphics}
\usepackage{subfigure}
\usepackage{wrapfig}
\usepackage{algorithm}
\usepackage{algorithmic}
\usepackage{threeparttable}
\usepackage{xcolor}
\setcitestyle{square}
\hypersetup{
colorlinks = true,
urlcolor=blue,
}

\setcitestyle{citesep={,}}

\newcommand{\xmark}{\ding{55}}%
\newcommand{\tabincell}[2]{\begin{tabular}{@{}#1@{}}#2\end{tabular}}

\newcommand{\hs}[1]{\hspace{#1ex}}

\title{\texttt{FedML}: A Research Library and Benchmark for Federated Machine Learning}

\author{
Chaoyang He\thanks{Corresponding authors. Email: \texttt{chaoyang.he@usc.edu}},\hs{3} Songze Li,\hs{3} Jinhyun So,\hs{3}
Xiao Zeng,\hs{3} Mi Zhang\\
\hs{3}USC \hs{8} Stanford \hs{7} USC\hs{10} MSU \hs{7} MSU\\

\bf \hs{1} Hongyi Wang,\hs{1} Xiaoyang Wang, \hs{1} Praneeth Vepakomma,\hs{1} Abhishek Singh,\hs{1} Hang Qiu  \\
\hs{-2} UW-Madison \hs{6} UIUC \hs{15}MIT \hs{15}MIT \hs{10}USC \\

\bf \hs{2}  Xinghua Zhu, Jianzong Wang, \hs{1} Li Shen, Peilin Zhao, \hs{3} Yan Kang, Yang Liu \\
\hs{5} Ping An Tech. \hs{14} Tencent \hs{16}WeBank \\

\bf Ramesh Raskar, \hs{1} Qiang Yang, \hs{1} Murali Annavaram\footnotemark[1], \hs{1} Salman Avestimehr\footnotemark[1]\\
\hs{0}MIT \hs{8}HKUST \hs{10} USC \hs{16} USC}

\begin{document}

\maketitle




\vspace{-5mm}
\begin{abstract}
Federated learning (FL) is a rapidly growing research field in machine learning. However, existing FL libraries cannot adequately support diverse algorithmic development; inconsistent dataset and model usage make fair algorithm comparison challenging.
In this work, we introduce \texttt{FedML}, an open research library and benchmark to facilitate FL algorithm development and fair performance comparison. \texttt{FedML} supports three computing  paradigms: on-device training for edge devices, distributed computing, and single-machine simulation. \texttt{FedML} also promotes diverse algorithmic research with flexible and generic API design and comprehensive reference baseline implementations (optimizer, models, and datasets).
We hope \texttt{FedML} could provide an efficient and reproducible means for developing and evaluating FL algorithms that would benefit the FL research community. We maintain the source code, documents, and user community at \url{https://fedml.ai}.

\end{abstract}

\section{Introduction}

\vspace{-3mm}
Federated learning (FL) is a distributed learning paradigm that aims to train machine learning models from scattered and isolated data \cite{kairouz2019advances}. FL differs from data center-based distributed training in three major aspects: 1) statistical heterogeneity, 2) system constraints, and 3) trustworthiness. Solving these unique challenges calls for efforts from a variety of fields, including machine learning, wireless communication, mobile computing, distributed systems, and information security, making federated learning a truly interdisciplinary research field.

In the past few years, more and more efforts have been made to address these unique challenges.
%
To tackle the challenge of statistical heterogeneity,  distributed optimization methods such as Adaptive Federated Optimizer \cite{reddi2020adaptive}, FedNova \cite{wang2020tackling}, FedProx \cite{Sahu2018OnTC}, and FedMA \cite{wang2020federated} have been proposed.
%
To tackle the challenge of system constraints, researchers apply sparsification and quantization techniques to reduce the communication overheads and computation costs during the training process \cite{lin2017deep,tang2018communication,tang2019texttt,philippenko2020artemis,amiri2020federated,haddadpour2020federated,tang2020communication}.
%
To tackle the challenge of trustworthiness, existing research focuses on developing new defense techniques for adversarial attacks to make FL robust \cite{hitaj2017deep,yin2018byzantine,zhu2019deep,nasr2019comprehensive,wang2019beyond,bhagoji2019analyzing,fung2018mitigating,bagdasaryan2020backdoor,wei2020framework,chen2020backdoor,sun2019can,enthoven2020overview,chen2020backdoor}, and proposing methods such as differential privacy (DP) and secure multiparty computation (SMPC) to protect privacy \cite{bonawitz2016practical,geyer2017differentially,orekondy2018gradient,ryffel2018generic,melis2019exploiting,truex2019hybrid,triastcyn2019federated,xu2019hybridalpha,triastcyn2020federated}.

Although a lot of progress has been made, existing efforts are confronted with a number of limitations that we argue are critical to FL research:

\vspace{-2mm}
\paragraph{Lack of support of diverse FL computing paradigms.} Distributed training libraries in PyTorch \cite{paszke2019pytorch}, TensorFlow \cite{abadi2016tensorflow}, MXNet \cite{chen2015mxnet}, and distributed training-specialized libraries such as Horovod \cite{sergeev2018horovod} and BytePS \cite{peng2019generic} are designed for distributed training in data centers.
Although simulation-oriented FL libraries such as TensorFlow-Federated (TFF) \cite{TFF2019}, PySyft \cite{ryffel2018generic}, and LEAF \cite{caldas2018leaf} are developed, they only support centralized topology-based FL algorithms like FedAvg \cite{mcmahan2017communication} or FedProx \cite{Sahu2018OnTC} with simulation in a single machine, making them unsuitable for FL algorithms which require the exchange of complex auxiliary information and customized training procedure.
Production-oriented libraries such as FATE \cite{yang2019federated} and PaddleFL \cite{ma2019paddlepaddle} are released by industry. However, they are not designed as flexible frameworks that aim to support algorithmic innovation for open FL problems. 

\vspace{-3mm}
\paragraph{Lack of support of diverse FL configurations.} 
FL is diverse in network topology, exchanged information, and training procedures. In terms of network topology, a variety of network topologies such as vertical FL \cite{hardy2017private,cheng2019secureboost,yang2019parallel,yang2019quasi,nock2018entity,mpmc-vertical-fl-2020,liu2020asymmetrically}, split learning \cite{gupta2018distributed,vepakomma2018split}, decentralized FL \cite{he2019central,lian2017can,ye2020multi,lalitha2019decentralized}, hierarchical FL \cite{wainakh2020enhancing,liao2019federated,briggs2020federated,abad2020hierarchical,luo2020hfel,liu2019client}, and meta FL \cite{jiang2019improving, khodak2019adaptive, fallah2020personalized} have been proposed. In terms of exchanged information, besides exchanging gradients and models, recent FL algorithms propose to exchange information such as pseudo labels in semi-supervised FL \cite{jeong2020federated} and architecture parameters in neural architecture search-based FL \cite{he2020fednas,singh2020differentially,xu2020neural}. In terms of training procedures, the training procedures in federated GAN \cite{hardy2019md,augenstein2019generative} and transfer learning-based FL \cite{Liu2020ASF,jeong2018communication,li2019fedmd,sharma2019secure,ahn2019wireless} 
are very different from the vanilla FedAvg algorithm \cite{mcmahan2017communication}. 
Unfortunately, such diversity in network topology, exchanged information, and training procedures is not supported in existing FL libraries.

\vspace{-3mm}
\paragraph{Lack of standardized FL algorithm implementations and benchmarks.} The diversity of libraries used for algorithm implementation in existing work makes it difficult to fairly compare their performance.
The diversity of benchmarks used in existing work also makes it difficult to fairly compare their performance.
The non-I.I.D. characteristic of FL makes such comparison even more challenging \cite{hsieh2019non}: training the same DNN on the same dataset with different non-I.I.D. distributions produces varying model accuracies; one algorithm that achieves higher accuracy on a specific non-I.I.D. distribution than the other algorithms may perform worse on another non-I.I.D. distribution. 
In Table \ref{tab:stat-datasets}, we summarize the datasets and models used in existing work published at the top tier machine learning conferences such as NeurIPS, ICLR, and ICML in the past two years. We observe that the experimental settings of these work differ in terms of datasets, non-I.I.D. distributions, models, and the number of clients involved in each round. Any difference in these settings could affect the results. 



\begin{table*}[t]
\centering
\caption{Comparison between \texttt{FedML} and existing federated learning libraries and benchmarks.}
\label{tab:comparison}
\resizebox{\linewidth}{!}{
\begin{threeparttable}
    \begin{tabular}{p{0.2\columnwidth}lcccccc}
    \toprule
    & & \textbf{TFF} & \textbf{FATE} & \textbf{PaddleFL} & \textbf{LEAF} & \textbf{PySyft} & \textbf{\colorbox{yellow!30}{FedML}}\\
    \midrule
\multirow{3}{0.2\columnwidth}{\textbf{Diversified  Computing  Paradigms}}& standalone simulation&\checkmark&\checkmark&\checkmark&\checkmark&\checkmark &\checkmark\\
\cmidrule(lr){2-8}
&distributed computing&\checkmark&\checkmark&\checkmark&\xmark&\checkmark &\checkmark\\
\cmidrule(lr){2-8}
& on-device training (Mobile, IoT)&\xmark&\xmark&\xmark&\xmark&\xmark &\checkmark\\
\midrule
 \multirow{3}{0.2\columnwidth}{\textbf{Flexible and Generic API Design
}} & topology customization & \xmark&\xmark&\xmark&\xmark&\checkmark &\checkmark\\
\cmidrule(lr){2-8}
&flexible message flow&\xmark&\xmark&\xmark&\xmark&\xmark &\checkmark\\
\cmidrule(lr){2-8}
&exchange message customization& \xmark&\xmark&\xmark&\xmark&\checkmark &\checkmark\\
\midrule
\multirow{4}{0.2\columnwidth}{\textbf{Standardized Algorithm Implementations}}&FedAvg&\checkmark&\checkmark&\checkmark&\checkmark&\checkmark &\checkmark\\
\cmidrule(lr){2-8}
&decentralized FL&\xmark&\xmark&\xmark&\xmark&\xmark &\checkmark\\
\cmidrule(lr){2-8}
&FedNAS (beyond gradient/model)&\xmark&\xmark&\xmark&\xmark&\xmark &\checkmark\\
\cmidrule(lr){2-8}
& VFL (vertical federated learning)&\xmark&\checkmark&\checkmark&\xmark&\xmark &\checkmark\\
\cmidrule(lr){2-8}
& SplitNN (split learning) &\xmark&\xmark&\checkmark&\xmark&\checkmark &\checkmark\\
\midrule
\multirow{4}{0.2\columnwidth}{\textbf{Standardized Benchmarks}}&linear models (e.g., Logistic Regression)&\checkmark&\checkmark&\checkmark&\checkmark&\checkmark &\checkmark\\
\cmidrule(lr){2-8}
&shallow NN (e.g., Bi-LSTM)&\checkmark&\checkmark&\checkmark&\checkmark&\checkmark &\checkmark\\
\cmidrule(lr){2-8}
&Model DNN (e.g., ResNet)&\xmark&\xmark&\xmark&\xmark&\xmark &\checkmark\\
\cmidrule(lr){2-8}
&vertical FL &\xmark&\checkmark&\xmark&\xmark&\xmark &\checkmark\\
       \bottomrule
    \end{tabular}
    \end{threeparttable}}
    \vspace{-4mm}
\end{table*}

In this work, we present \texttt{FedML}, an open research library and benchmark to address the aforementioned limitations and facilitate FL research. \texttt{FedML} provides an end-to-end toolkit to facilitate FL algorithm development and fair performance comparison under diverse computing paradigms and configurations.
Table \ref{tab:comparison} summarizes the key differences between \texttt{FedML} and existing FL libraries and benchmarks. The highlights of \texttt{FedML} are summarized below:

\textbf{(i) Support of diverse FL computing paradigms}. \texttt{FedML} supports three diverse computing paradigms: 1) on-device training for edge devices including smartphones and Internet of Things (IoT), 2) distributed computing, and 3) single-machine simulation to meet algorithmic and system-level research requirements under different system deployment scenarios. 

\textbf{(ii) Support of diverse FL configurations}. \texttt{FedML} introduces a worker/client-oriented programming interface to enable diverse network topologies, flexible information exchange among workers/clients, and various training procedures. 

\textbf{(iii) Standardized FL algorithm implementations}. \texttt{FedML} includes standardized implementations of status quo FL algorithms. These implementations not only help users to familiarize \texttt{FedML} APIs but also can be used as baselines for comparisons with newly developed FL algorithms.

\textbf{(iv) Standardized FL benchmarks}. \texttt{FedML} provides standardized benchmarks with well-defined evaluation metrics, multiple synthetic and real-world non-I.I.D. datasets, as well as verified baseline results to facilitate fair performance comparison. 

\vspace{1mm}
\textbf{(v) Fully open and evolving}.
FL is a research field that evolves at a considerably fast pace.
%
This requires \texttt{FedML} to adapt at the same pace. We will continuously expand \texttt{FedML} to optimize three computing paradigms and support more algorithms (distributed optimizer) and benchmarks (models and datasets) for newly explored usage scenarios. \texttt{FedML} is fully open and welcomes contributions from the FL research community as well.
We hope researchers in diverse FL applications could contribute more valuable models and 
realistic datasets to our community. 
Promising application domains include, but are not limited to, computer vision \cite{hsu2020federated,liu2020fedvision}, natural language processing \cite{hard2018federated,leroy2019federated,ge2020fedner,chen2019federated,liu2020federated}, finance \cite{yang2019federated,cheng2019secureboost,liu2020fedcoin}, transportation \cite{elbir2020federated,lim2020towards,saputra2020federated,liu2020privacy,mirshghallah2020privacy,yin2020fedloc,chen2020practical,liang2019federated,saputra2019energy,anastasiou2019admsv2}, digital health \cite{rieke2020future,liu2018fadl,sheller2018multi,ju2020privacy,ju2020federated,li2019privacy,chen2020fedhealth}, recommendation \cite{flanagan2020federated,chen2020robust,li2020federated,qi2020fedrec,ribero2020federating,ammad2019federated}, robotics \cite{liu2019federated,liu2019lifelong}, and smart cities \cite{wang2020cloud,albaseer2020exploiting}.

\begin{figure}[!ht]
\centering
{\includegraphics[width=0.9\textwidth]{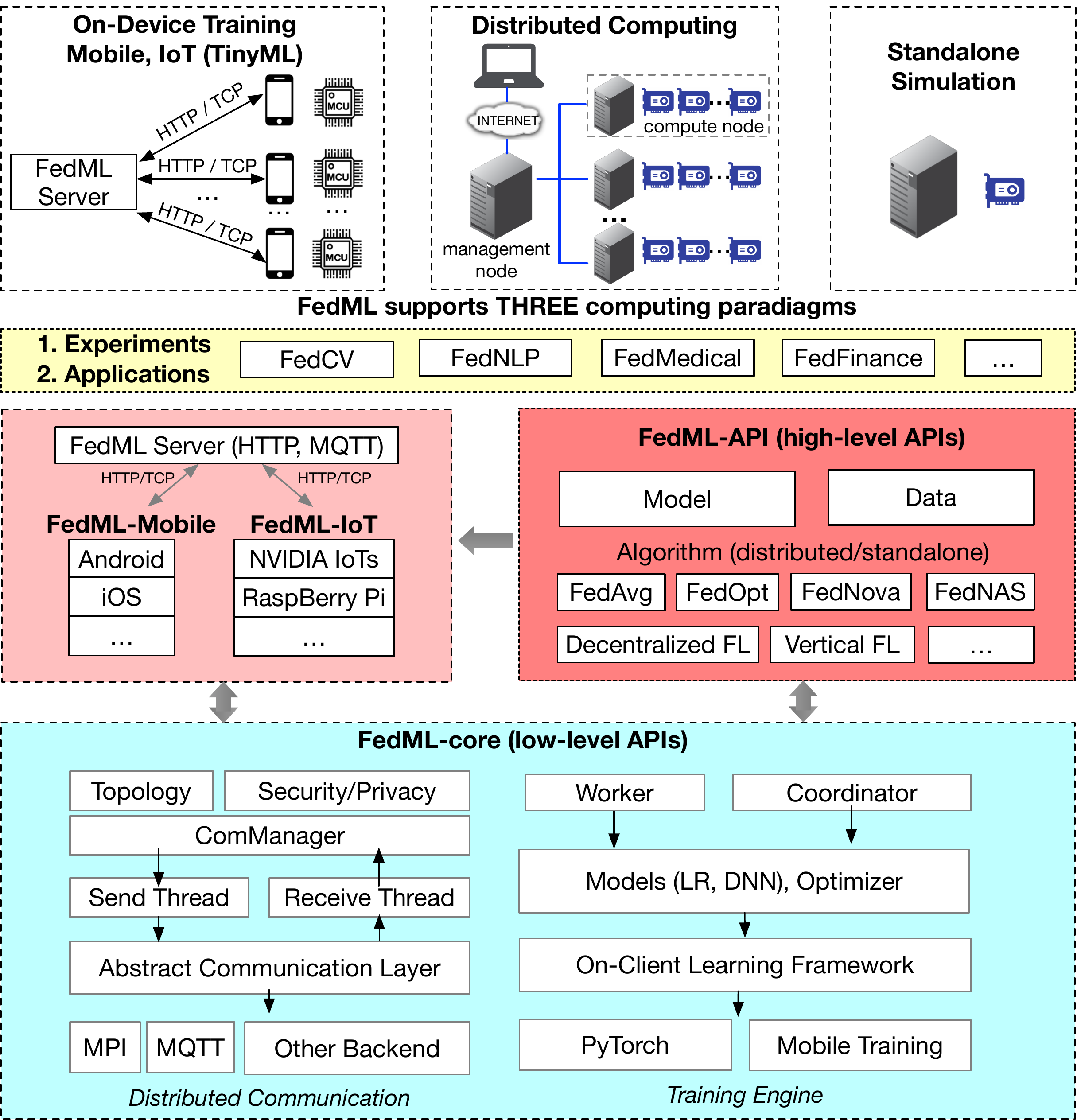}}
\caption{\textcolor{black}{Overview of \texttt{FedML} library.}}
\label{fig:system}
\end{figure}

\section{\texttt{FedML} Library: Architecture Design}
\label{sec:arch}


Figure \ref{fig:system} provides an overview of \texttt{FedML} library. 
The \texttt{FedML} library has two key components: \texttt{FedML-API} and \texttt{FedML-core}, which represents high-level API and low-level API, respectively. 

\texttt{FedML-core} separates distributed communication and model training into two separate modules. The distributed communication module is responsible for low-level communication among different workers/clients. The communication backend is based on MPI (message passing interface)\footnote{\url{https://pypi.org/project/mpi4py/}}. 
Inside the distributed communication module, a \texttt{TopologyManager} supports a variety of network topologies that can be used in many existing FL algorithms \cite{he2019central,lian2017can,ye2020multi,lalitha2019decentralized}. In addition,  security/privacy-related functions are also supported.
The model training module is built upon PyTorch. Users can implement workers (trainers) and coordinators according to their needs. 

\texttt{FedML-API} is built upon \texttt{FedML-core}. With the help of \texttt{FedML-core}, new algorithms in distributed version can be easily implemented by adopting the client-oriented programming interface, which is a novel design pattern for flexible distributed computing (Section \ref{sec:pi}). Such a distributed computing paradigm is essential for scenarios in which large DNN training cannot be handled by standalone simulation due to GPU memory and training time constraints. This distributed computing design is not only used for FL, but it can also be used for conventional in-cluster large-scale distributed training. \texttt{FedML-API} also suggests a machine learning system practice that separates the implementations of models, datasets, and algorithms. This practice enables code reuse and fair comparison, avoiding statistical or system-level gaps among algorithms led by non-trivial implementation differences. Another benefit is that FL applications can develop more models and submit more realistic datasets without the need to understand the details of different distributed optimization algorithms.

One key feature of \texttt{FedML} is its support of FL on real-world hardware platforms.
Specifically, \texttt{FedML} includes \texttt{FedML-Mobile} and \texttt{FedML-IoT}, which are two on-device FL testbeds built upon real-world hardware platforms. 
Currently, \texttt{FedML-Mobile} supports Android smartphones and \texttt{FedML-IoT} supports Raspberry PI 4 and NVIDIA Jetson Nano (see Appendix \ref{iot_devices} for details). 
With such testbeds built upon real-world hardware platforms, researchers can evaluate realistic system performance, such as training time, communication, and computation cost. 
To support conducting experiments on those real-world hardware platforms, our \texttt{FedML} architecture design can smoothly transplant the distributed computing code to the \texttt{FedML-Mobile} and \texttt{FedML-IoT} platforms, reusing nearly all algorithmic implementations in the distributed computing paradigm. Moreover, for \texttt{FedML-IoT}, researchers only need to program with Python to customize their research experiments without the need to learn new system frameworks or programming languages (e.g., Java, C/C++)\footnote{Please check here for details: \url{https://github.com/FedML-AI/FedML/tree/master/fedml_iot}}.

\section{\texttt{FedML} Library: Programming Interface}
\label{sec:pi}


The goal of the \texttt{FedML} programming interface is to provide simple user experience to allow users to build distributed training applications (e.g. design customized message flow and topology definitions) by only focusing on algorithmic implementations while ignoring the low-level communication backend details.

\begin{figure}[!ht]
\centering
\subfigure[\label{fig:overview_training_oriented} Training procedure-oriented programming ]{{\includegraphics[width=0.495\textwidth]{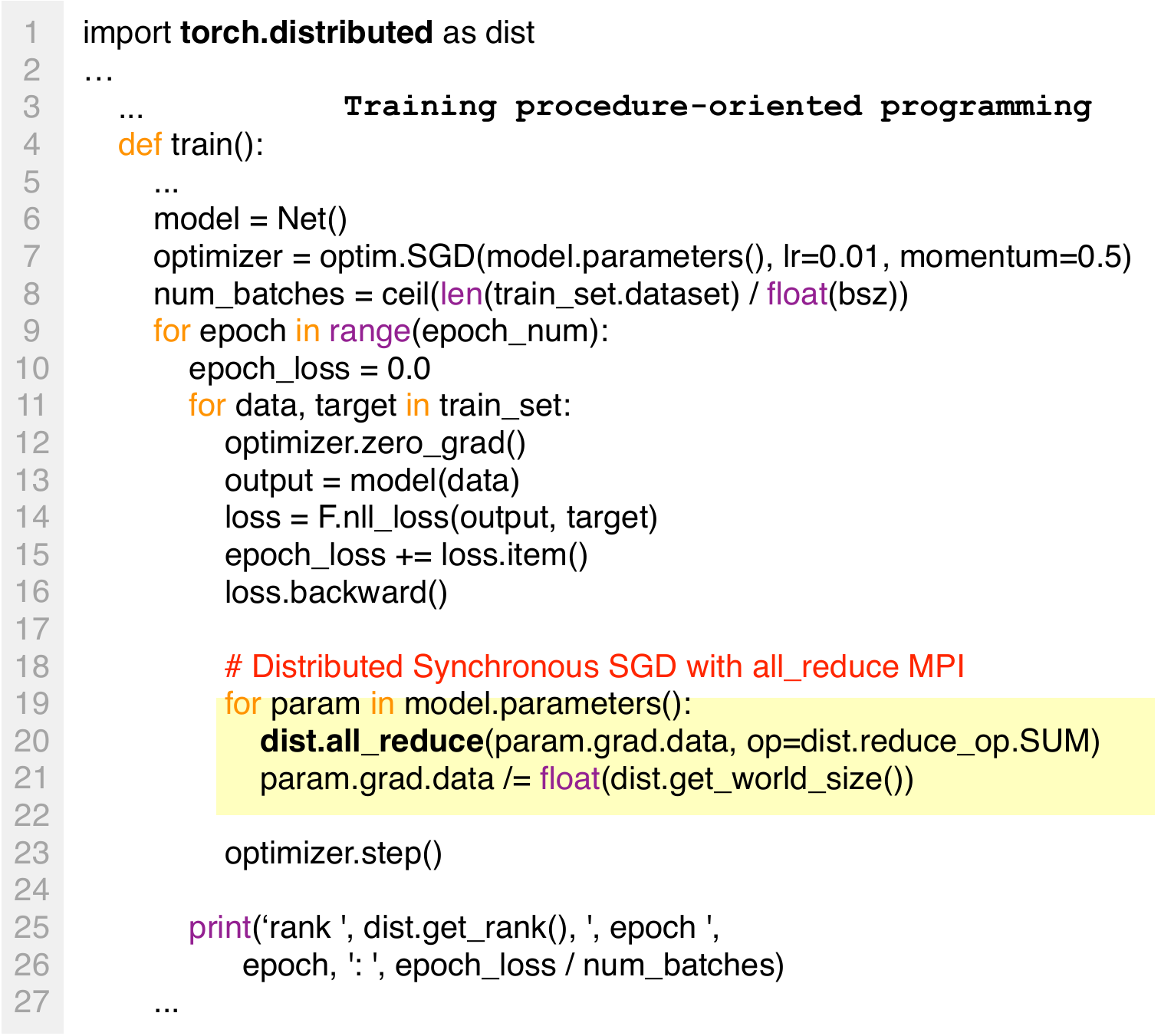}}}
\subfigure[\label{fig:overview_worker_oriented} Worker-oriented programming ]{{\includegraphics[width=0.495\textwidth]{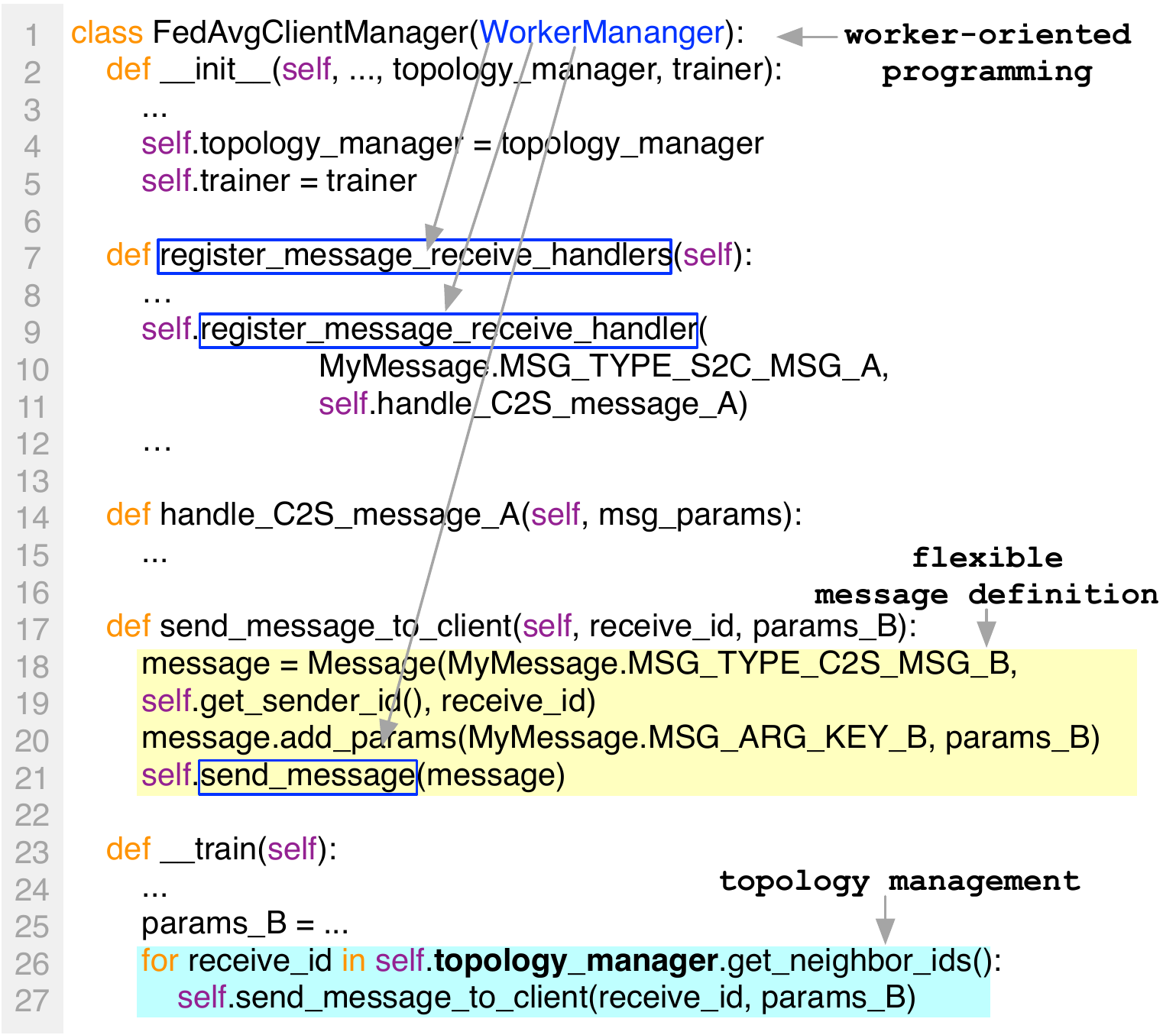}}}
\vspace{-2mm}
\caption{\textcolor{black}{A worker-oriented programming design pattern of \texttt{FedML}}.
\vspace{-2mm}}
\label{fig:overview}
\end{figure}

\paragraph{Worker/client-oriented programming.} As shown in Figure \ref{fig:overview_worker_oriented}, \texttt{FedML} provides the worker-oriented programming design pattern, which can be used to program the worker behavior when participating in training or coordination in the FL algorithm. We describe it as worker-oriented because its counterpart, the standard distributed training library (as the \texttt{torch.distributed} example\footnote{More details can be found at \url{https://pytorch.org/tutorials/intermediate/dist_tuto.html}} shown in Figure \ref{fig:overview_training_oriented}), normally completes distributed training programming by describing the entire training procedure rather than focusing on the behavior of each worker. 

With the worker-oriented programming design pattern, the user can customize its own worker in FL network by inheriting the \textcolor{blue}{\texttt{WorkerManager}} class and utilizing its predefined APIs \textcolor{blue}{\texttt{register\_message\_receive\_handler}} and \textcolor{blue}{\texttt{send\_message}} to define the receiving and sending messages without considering the underlying communication mechanism (as shown in the highlighted \textcolor{blue}{blue box} in Figure \ref{fig:overview_worker_oriented}). Conversely, existing distributed training frameworks do not have such flexibility. In order to make the comparison clearer, we use the most popular machine learning framework PyTorch as an example. Figure \ref{fig:overview_training_oriented} illustrates a complete training procedure (distributed synchronous SGD) and aggregates gradients from all other workers with the \texttt{all\_reduce} messaging passing interface. Although it supports multiprocessing training, it cannot flexibly customize different messaging flows in any network topology. In PyTorch, another distributed training API, \texttt{torch.nn.parallel.paraDistributedDataParallel}\footnote{It is recommended to use \texttt{torch.nn.parallel.paraDistributedDataParallel} instead of \texttt{torch.nn.DataParallel}. For more details, please refer to \url{https://pytorch.org/tutorials/intermediate/ddp\_tutorial.html} and \url{https://pytorch.org/docs/master/notes/cuda.html\#cuda-nn-ddp-instead}}, also has such inflexibly. 

\paragraph{Message definition beyond gradient and model.} \texttt{FedML} also supports message exchange beyond the gradient or model from the perspective of message flow. This type of auxiliary information may be due to either the need for algorithm design or the need for system-wide configuration delivery. Each worker defines the message type from the perspective of sending. Thus, in the above introduced worker-oriented programming, the \textcolor{blue}{\texttt{WorkerManager}} should handle messages defined by other trainers and also send messages defined by itself. The sending message is normally executed after handling the received message. As shown in Figure \ref{fig:overview_worker_oriented}, \colorbox{yellow!30}{in the yellow background highlighted code snippet}, workers can send any message type and related message parameters using the \texttt{train()} function.
\label{sec:topologymanagement}
\begin{figure}[!ht]
    \centering
    \includegraphics[width=\textwidth]{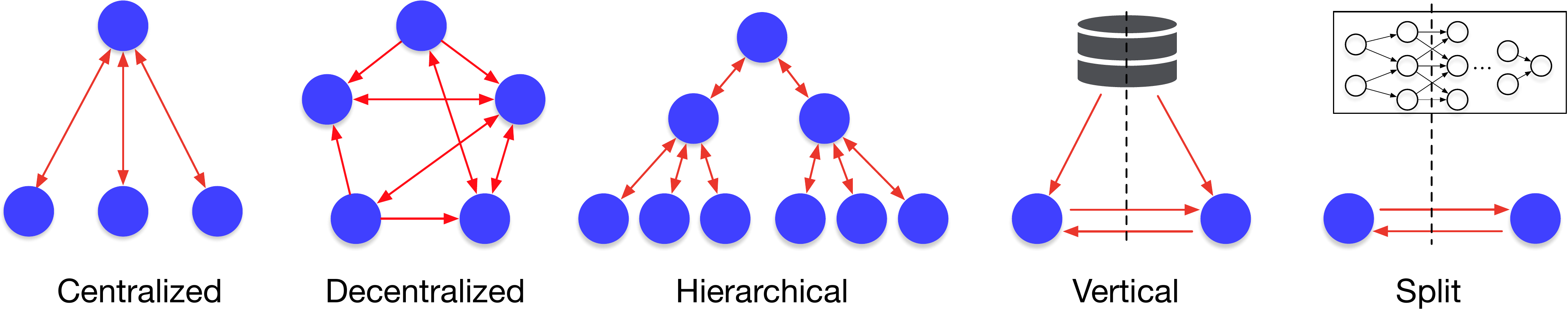}
    \vspace{-3mm}
    \caption{Illustration of various topology definitions in federated learning.}
    \vspace{-2mm}
    \label{fig:topology}
\end{figure}

\vspace{-4mm}
\paragraph{Topology management.} As demonstrated in Figure \ref{fig:topology}, FL has various topology definitions, such as vertical FL \cite{hardy2017private,cheng2019secureboost,yang2019parallel,yang2019quasi,nock2018entity,mpmc-vertical-fl-2020,liu2020asymmetrically}, split learning \cite{gupta2018distributed,vepakomma2018split}, decentralized FL \cite{he2019central,lian2017can,ye2020multi,lalitha2019decentralized}, and Hierarchical FL \cite{wainakh2020enhancing,liao2019federated,briggs2020federated,abad2020hierarchical,luo2020hfel,liu2019client}. 
In order to meet such diverse requirements, \texttt{FedML} provides \texttt{TopologyManager} to manage the topology and allows users to send messages to arbitrary neighbors during training. Specifically, after the initial setting of \texttt{TopologyManager} is completed, for each trainer in the network, the neighborhood worker ID can be queried via the \texttt{TopologyManager}. 
\colorbox{cyan!30}{In line \textcolor{red}{26} of Figure \ref{fig:overview_worker_oriented}}, we see that the trainer can query its neighbor nodes through the \texttt{TopologyManager} before sending its message. 

\paragraph{Trainer and coordinator.} We also need the coordinator to complete the training (e.g., in FedAvg, the central worker is the coordinator while the others are trainers). For the trainer and coordinator, \texttt{FedML} does not over-design. Rather, it gives the implementation completely to the developers, reflecting the flexibility of our framework. The implementation of the trainer and coordinator is similar to the process in Figure \ref{fig:overview_training_oriented}, which is consistent with the training implementation of a standalone version training. We provide some reference implementations of different trainers and coordinators in our source code (Section \ref{sec:reference_examples}).

\paragraph{Privacy, security, and robustness.}
While the FL framework facilitates data privacy \cite{mirshghallah2020privacy} by keeping data locally available to the users and only requiring communication for model updates, users may still be concerned about partial leakage of their data which may be inferred from the communicated model (e.g.,~\cite{fredrikson2015model}). Aside from protecting the privacy of users' data, another critical security requirement for the FL platform, especially when operating over mobile devices, is the robustness towards user dropouts. Specifically, to accomplish the aforementioned goals of achieving security, privacy, and robustness, various cryptography and coding-theoretic approaches have been proposed to manipulate intermediate model data~ \cite{bonawitz2017practical,so2020turbo}.

To facilitate rapid implementation and evaluation of data manipulation techniques to enhance security, privacy, and robustness, we include low-level APIs that implement common cryptographic primitives such as secrete sharing, key agreement, digital signature, and public key infrastructure. We also plan to include an implementation of \texttt{Lagrange Coded Computing} (LCC)~\cite{yu2019lagrange}. LCC is a recently developed coding technique on data that achieves optimal resiliency, security (against adversarial nodes), and privacy for any polynomial evaluations on the data. Finally, we plan to provide a sample implementation of the secure aggregation algorithm~\cite{bonawitz2017practical} using the above APIs.

In standard FL settings, it is assumed that there is no single central authority that owns or verifies the training data or user hardware, and it has been argued by many recent studies that FL lends itself to new adversarial attacks during decentralized model training~\cite{bagdasaryan2020backdoor,sun2019can,bhagoji2019analyzing,xie2019dba,wang2020attack}. Several robust aggregation methods have been proposed to enhance the robustness of FL against adversaries ~\cite{sun2019can,pillutla2019robust,blanchard2017machine}.

To accelerate generating benchmark results on new types of adversarial attacks in FL, we include the latest robust aggregation methods presented in literature including (i) norm difference clipping~\cite{sun2019can}; weak differential private (DP)~\cite{sun2019can}; (ii) RFA (geometric median)~\cite{pillutla2019robust}; (iii) \textsc{Krum} and (iv) \textsc{Multi-Krum}~\cite{blanchard2017machine}. Our APIs are easily extendable to support newly developed types of robust aggregation methods. On the attack end, we observe that most of the existing attacks are highly task-specific. Thus, it is challenging to provide general adversarial attack APIs. Our APIs support the backdoor with model replacement attack presented in~\cite{bagdasaryan2020backdoor} and the edge-case backdoor attack presented in~\cite{wang2020attack} to provide a reference for researchers to develop new attacks.

\vspace{-2mm}
\section{\texttt{FedML} Benchmark: Algorithms, Models, and Datasets}
\vspace{-2mm}
\label{sec:benchmark}

\subsection{Algorithms: Federated Optimizer}
\label{sec:reference_examples}

\begin{figure}[!ht]
\centering
\subfigure[\label{fig:example_fedavg} Centralized FL ]{{\includegraphics[width=0.21\textwidth]{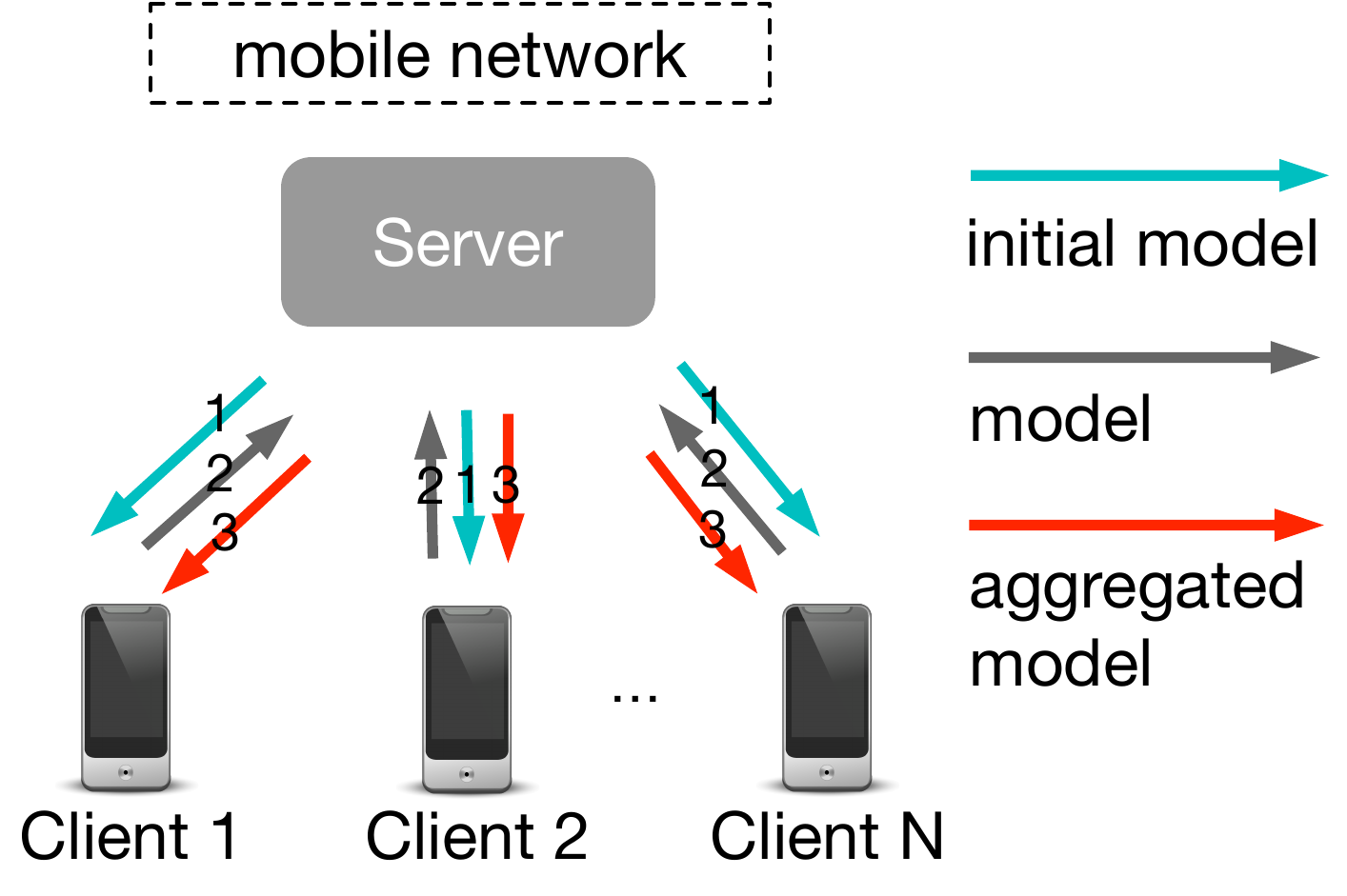}}}
\hfill
\subfigure[\label{fig:example_decentralized_fl} Decentralized FL]{{\includegraphics[width=0.22\textwidth]{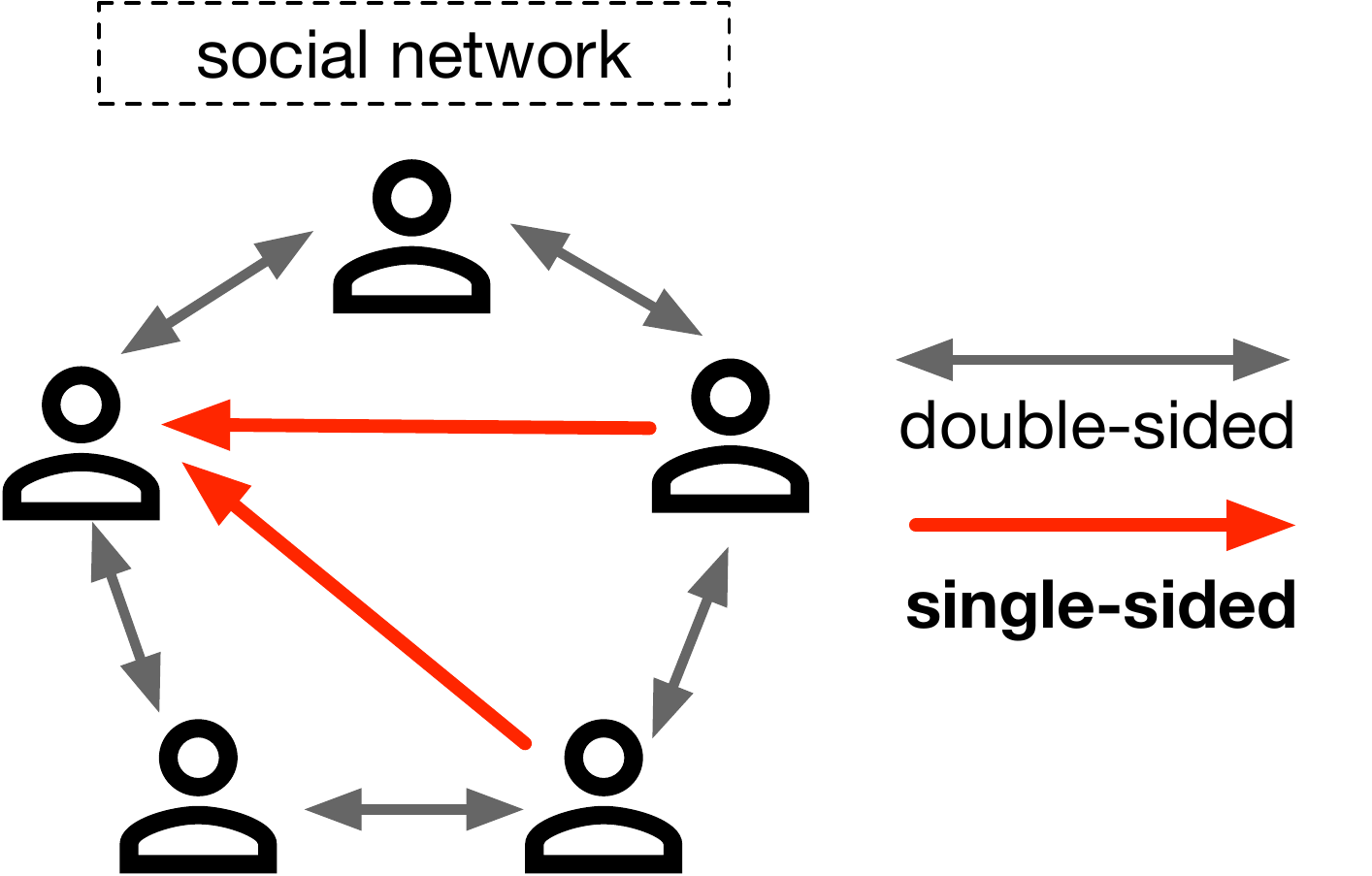}}}
 \hfill
\subfigure[\label{fig:example_vfl} Vertical FL ]{{\includegraphics[width=0.215\textwidth]{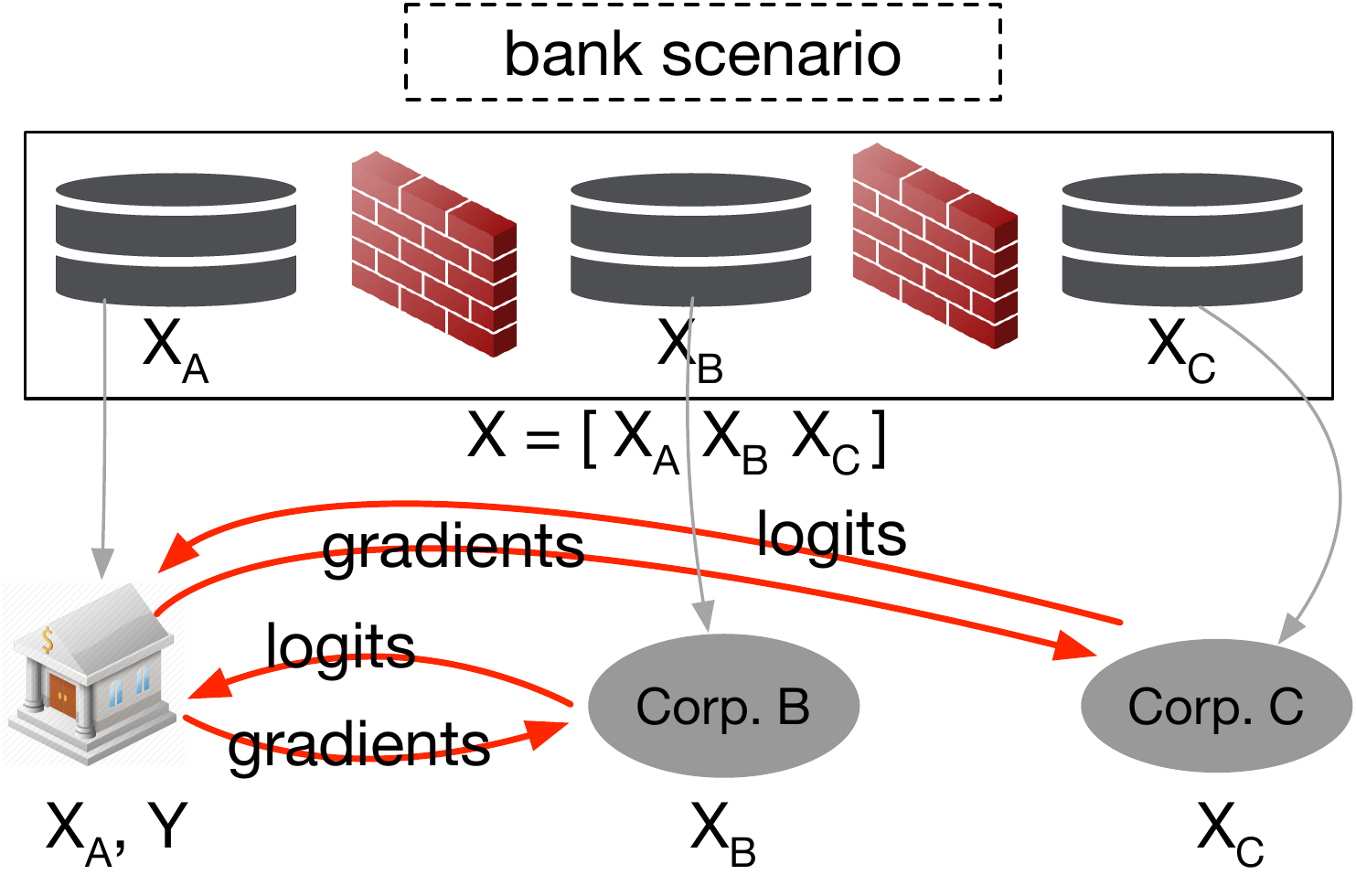}}}\hfill
\subfigure[\label{fig:example_sl} FedNAS ]{{\includegraphics[width=0.255\textwidth]{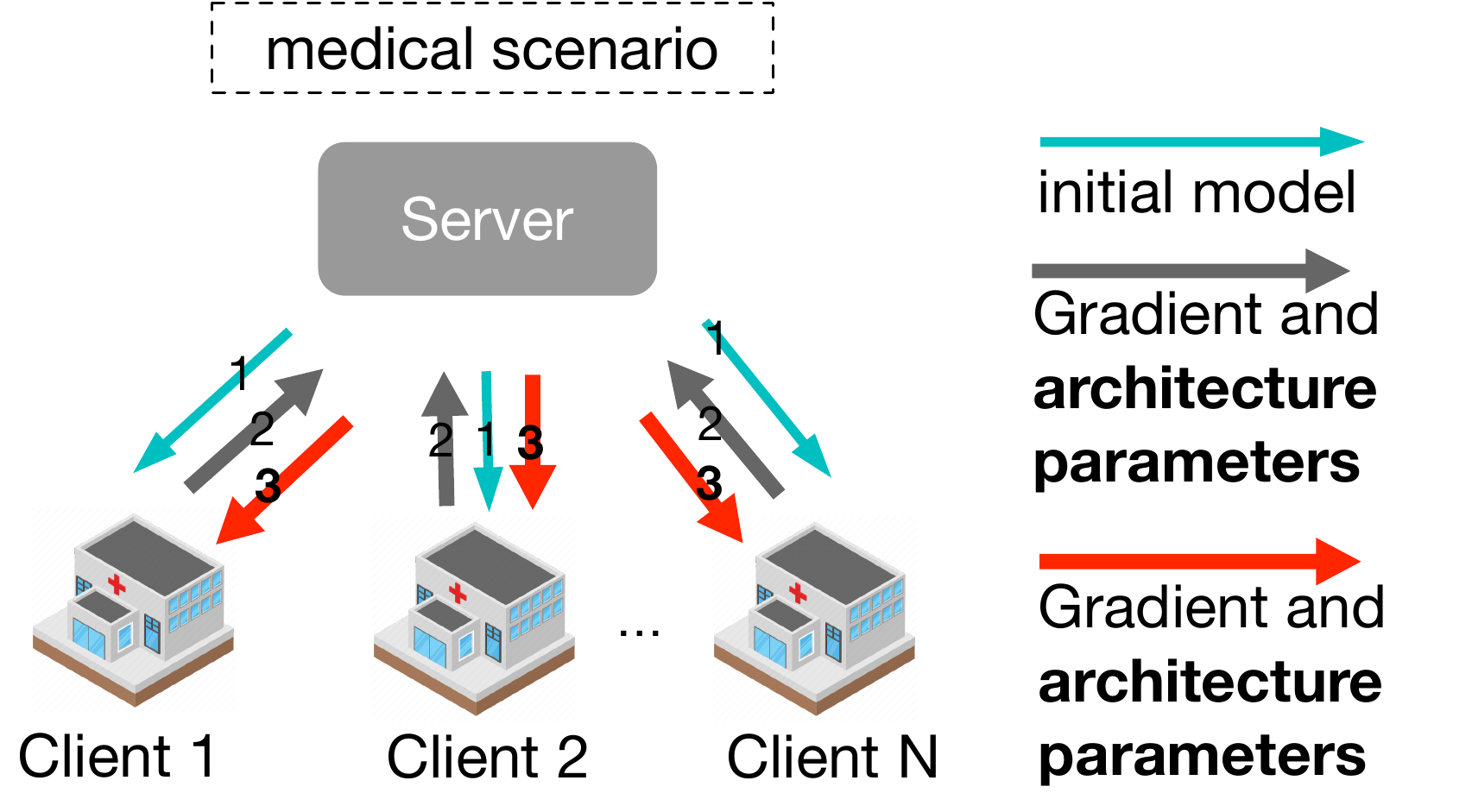}}}
\caption{\textcolor{black}{Supported algorithms that are diverse in network topology, exchanged information, and training procedures.}}
\label{fig:examples}
\end{figure}
As shown in Figure \ref{fig:examples}, \texttt{FedML} is capable of supporting FL algorithms that are diverse in network topology, exchanged information, and training procedures. These supported algorithms can be used as implementation examples and baselines to help users develop and evaluate their own algorithms. Currently, \texttt{FedML} includes the standard implementations of multiple status quo FL algorithms: Federated Averaging (FedAvg) \cite{mcmahan2017communication}, Decentralized FL \cite{he2019central}, Vertical Federated Learning (VFL) \cite{fmlyangqiang}, Split learning \cite{gupta2018distributed, vepakomma2018split}, Federated Neural Architecture Search (FedNAS) \cite{he2020fednas}, and Turbo-Aggregate \cite{so2020turbo}. Fore more details of these algorithms, please refer to Appendix \ref{appendix:alg}. 

We will keep following the latest algorithm to be published at top-tier machine learning conferences, and will continuously add new FL algorithms such as Adaptive Federated Optimizer \cite{reddi2020adaptive}, FedNova \cite{wang2020tackling}, FedProx \cite{Sahu2018OnTC}, and FedMA \cite{wang2020federated} in near future.




\subsection{Models and Datasets}
Inconsistent usage of datasets, models, and non-I.I.D. partition methods makes it difficult to fairly compare the performance of FL algorithms (in Table \ref{tab:stat-datasets}, we summarize the non-I.I.D. datasets and models used in existing work published at the top tier  machine learning venues in the past two years). To enforce fair comparison, \texttt{FedML} benchmark explicitly specifies the combinations of datasets, models, and non-I.I.D. partition methods to be used for experiments. In particular, we divide the benchmark into three categories: 1) linear models (convex optimization),  2) lightweight shallow neural networks (non-convex optimization), and 3) deep neural networks (non-convex optimization).

\vspace{-2mm}
\begin{table*}[!ht]
\centering
\caption{Federated datasets for linear models (convex optimization).}
\label{tab:benchmark-convex}
\resizebox{\linewidth}{!}{
\begin{threeparttable}
    \begin{tabular}{lcccccc}
    \toprule
      \textbf{Datasets} & \textbf{\# of training} & \textbf{\# of testing} & \textbf{non-I.I.D. }& \textbf{\# of clients /}&  \textbf{baseline model}\\
      &\textbf{samples} & \textbf{samples}  &\textbf{partition method} &\textbf{devices} &\\
      \midrule
      MNIST &60000 & 10000&power law & 1000& logistic regression\\
      Federated EMNIST &671585 & 77483&realistic partition& 3400& logistic regression\\
      Synthetic $(\alpha, \beta)$ \cite{li2018federated} & 4305 & 4672 & refer to \cite{li2018federated} & 30 & logistic regression\\
      
       \bottomrule
    \end{tabular}
\end{threeparttable}}
\end{table*}

\paragraph{Federated datasets for linear models (convex optimization).} The linear model category is used for convex optimization experiments such as the ones in \cite{li2018federated} and \cite{li2019convergence}. In this category, we include three datasets (Table \ref{tab:benchmark-convex}): MNIST \cite{lecun1998gradient}, Federated EMNIST \cite{reddi2020adaptive}, and Synthetic ($\alpha$, $\beta$) \cite{li2018federated}, with the logistic regression as the baseline model.

\vspace{-2mm}
\begin{table*}[!ht]
\centering
\caption{Federated datasets for lightweight shallow neural networks (non-convex optimization).}
\label{tb:benchmark-non-convex}
\resizebox{\linewidth}{!}{
\begin{threeparttable}
    \begin{tabular}{lcccccc}
    \toprule
      \textbf{Datasets} & \textbf{\# of training} & \textbf{\# of testing} & \textbf{partition method}& \textbf{\# of clients /}&  \textbf{baseline model}\\
      &\textbf{samples} & \textbf{samples}  & &\textbf{devices} &\\
      \midrule
      Federated EMNIST &671585 & 77483&realistic partition& 3400& CNN (2 Conv + 2 FC)\cite{reddi2020adaptive}\\
      CIFAR-100& 50000& 10000& Pachinko Allocation & 500&  ResNet-18 + group normalization\\
      Shakespeare& 16068& 2356&realistic partition & 715&  RNN (2 LSTM + 1 FC)\\
      StackOverflow& 135818730& 16586035& realistic partition&342477 & RNN (1 LSTM + 2 FC)\\
       \bottomrule
    \end{tabular}
\end{threeparttable}}
\end{table*}

\paragraph{Federated datasets for lightweight shallow neural networks (non-convex optimization).} Due to resource constraints of edge devices, shallow neural networks are commonly used in existing work for experiments. In this category,  we include four datasets (Table \ref{tb:benchmark-non-convex}):  Federated EMNIST \cite{caldas2018leaf}, CIFAR-100 \cite{krizhevsky2009learning}, Shakespeare \cite{mcmahan2017communication}, and StackOverflow \cite{TTFstackoverflow}. Please refer to Appendix \ref{appendix:dataset} for more details.

\vspace{-2mm}
\begin{table*}[!ht]
\centering
 \caption{Federated datasets for deep neural networks.}
\label{tab:benchmark-deep}
\resizebox{\linewidth}{!}{
\begin{threeparttable}
    \begin{tabular}{lcccccc}
    \toprule
      \textbf{Datasets} & \textbf{\# of training} & \textbf{\# of testing} & \textbf{partition method}& \textbf{\# of clients /}& \textbf{baseline model}\\
      &\textbf{samples} & \textbf{samples}  & &\textbf{devices} &\\
      \midrule
      CIFAR-10 &50,000 & 10,000& Latent Dirichlet Allocation& 10 & ResNet-56, MobileNet\\
      CIFAR-100& 50,000& 10,000& Latent Dirichlet Allocation& 10&  ResNet-56, MobileNet\\
      CINIC-10 & 90,000& 90,000& Latent Dirichlet Allocation & 10&  ResNet-56, MobileNet\\
StackOverflow& 135,818,730& 10,586,035& realistic partition&342477 (10)  & RNN (2 LSTM + 1 FC)\\
       \bottomrule
    \end{tabular}
\end{threeparttable}}
\end{table*}

\paragraph{Federated datasets for deep neural networks (non-convex optimization).} Given the resource constraints of edge devices, large DNN models are usually trained under the cross-organization FL (also called cross-silo FL) setting. For example, \cite{he2020fednas} has studied large DNN models for cross-silo FL in the hospital scenario. However, large DNN models dominate the accuracy in most learning tasks. Pushing FL of large DNN models on edge devices is challenging but a meaningful endeavor, which motivates us to make this benchmark category. For example, \cite{he2020group} has proposed an efficient training algorithm for large CNN models on edge devices. Table \ref{tab:benchmark-deep} shows datasets and models we include in this category. Please refer to Appendix \ref{appendix:dataset} for more details. 



\vspace{-1mm}
\section{Experiments}
\vspace{-1mm}
\texttt{FedML} provides benchmark experimental results as references for newly developed FL algorithms. To ensure real-time updates, we maintain benchmark experimental results using \texttt{Weight and Bias}\footnote{\url{https://www.wandb.com/}}. The web link to view all the benchmark experimental results can be found at our GitHub repository.


\begin{figure}[h!]
\centering
\setcounter{subfigure}{0}
\subfigure[\label{fig:training_curve1} ResNet-56 on CIFAR-10]
{{\includegraphics[width=0.3\textwidth]{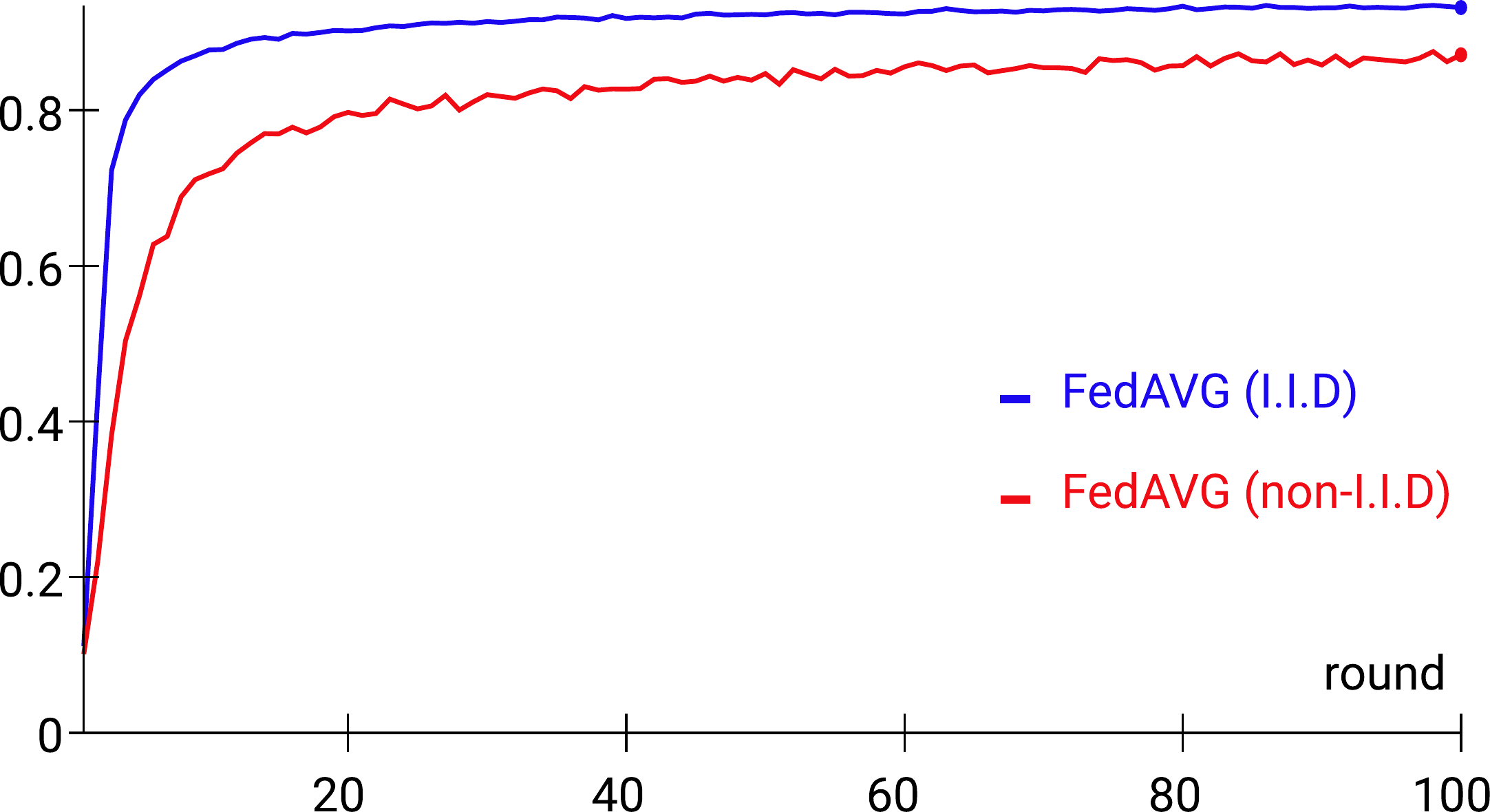}}}
\hfill
\subfigure[\label{fig:training_curve2} ResNet-56 on CIFAR-100]
{{\includegraphics[width=0.3\textwidth,height=2.3cm]{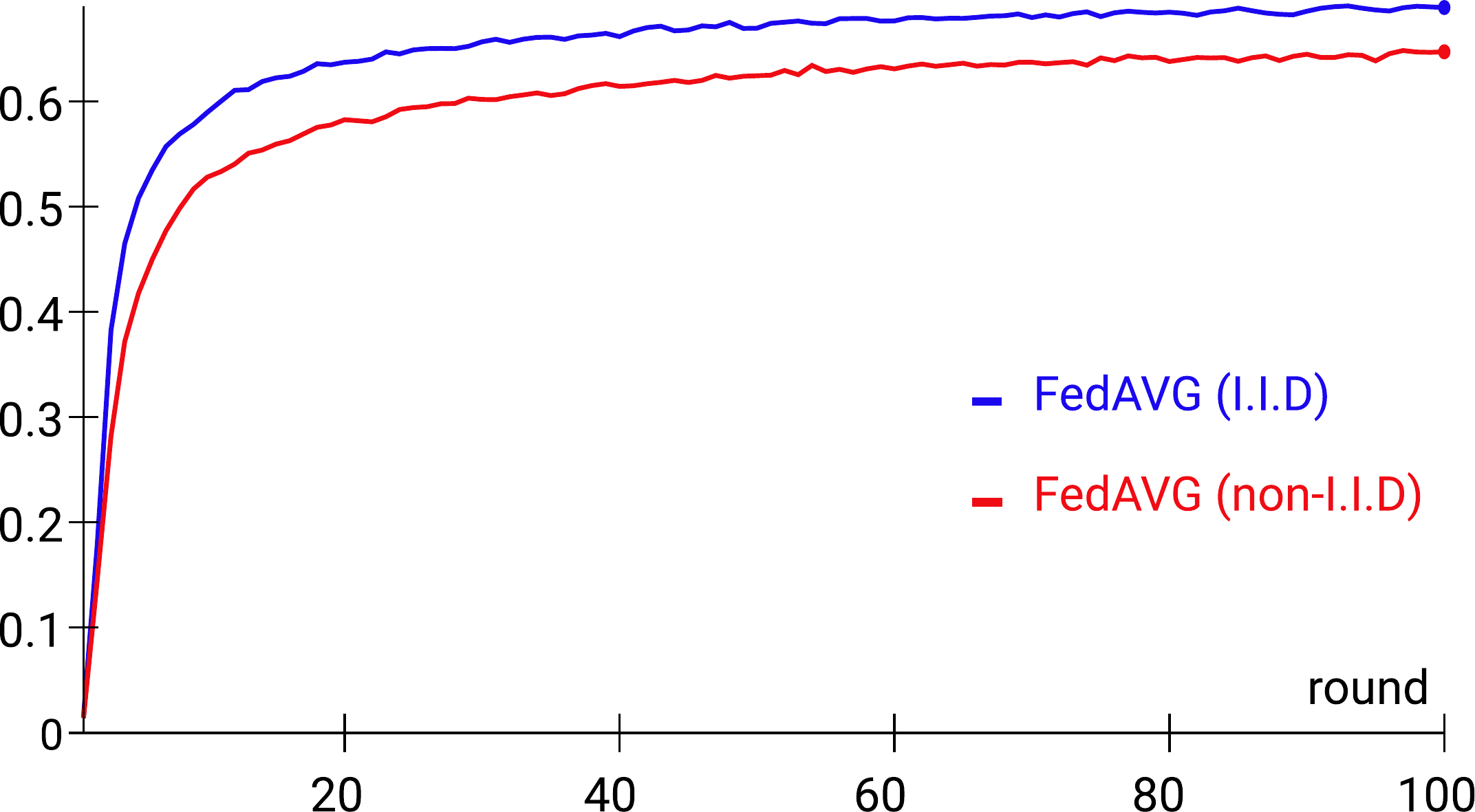}}}
\hfill
\subfigure[\label{fig:training_curve3} ResNet-56 on CINIC-10]
{{\includegraphics[width=0.3\textwidth,height=2.3cm]{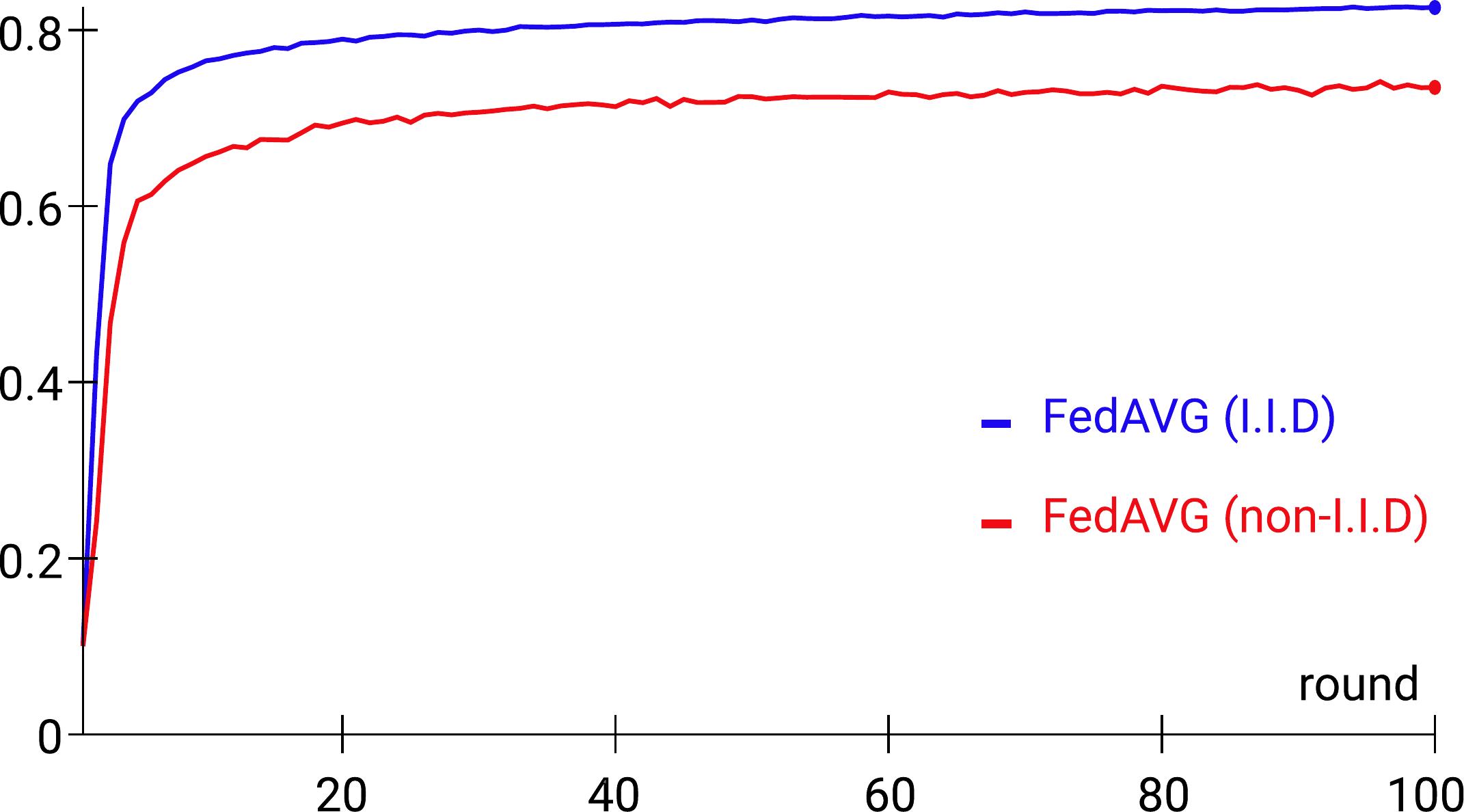}}}
\caption{\textcolor{black}{Test accuracy of ResNet-56 during training.}}
\vspace{-2mm}
\label{fig:training_curve}
\end{figure}

\vspace{-3mm}
\begin{table*}[!ht]
\centering
 \caption{Experimental results of training modern CNNs.}
\label{tab:ex-results}
\resizebox{\linewidth}{!}{
\begin{threeparttable}
    \begin{tabular}{lcccccccc}
    \toprule
      \textbf{Dataset} & \textbf{Non-I.I.D.} & \textbf{Model} & \textbf{Number of Workers}& \textbf{Algorithm}& \textbf{Acc. on}& \textbf{Acc. on}\\
      &\textbf{Partition Method} &  & &  &\textbf{I.I.D} &\textbf{non-I.I.D}\\
      \midrule
      \multirow{2}{*}{CIFAR-10}& \multirow{2}{*}{Latent Dirichlet Allocation} &ResNet-56 &\multirow{2}{*}{10} &FedAvg &93.19 &87.12 ($\downarrow \textcolor{red}{6.07}$)\\
      & &MobileNet & &FedAvg& 91.12 & 86.32 ($\downarrow  \textcolor{red}{4.80}$)\\
      \midrule
      \multirow{2}{*}{CIFAR-100}&\multirow{2}{*}{Latent Dirichlet Allocation} &ResNet-56 &\multirow{2}{*}{10} &FedAvg &68.91 &64.70 ($\downarrow \textcolor{red}{4.21}$)\\
      & &MobileNet &&FedAvg&55.12  &53.54 ($\downarrow  \textcolor{red}{1.58}$)\\
      \midrule
      \multirow{2}{*}{CINIC-10}&\multirow{2}{*}{Latent Dirichlet Allocation} &ResNet-56 &\multirow{2}{*}{10} &FedAvg & 82.57&73.49 ($\downarrow\textcolor{red}{9.08})$\\
      & &MobileNet & &FedAvg&79.95&71.23 ($\downarrow  \textcolor{red}{8.72}$)\\
       \bottomrule
    \end{tabular}
    \begin{tablenotes}[para,flushleft]
      \footnotesize
      \item *Note: to reproduce the result, please use the same random seeds we set in the library.
    \end{tablenotes}
\end{threeparttable}}
\end{table*}

To demonstrate the capability of FedML, we ran experiments in a real distributed computing environment. We trained two CNNs (ResNet-56 and MobileNet) using the standard FedAvg algorithm. Table \ref{tab:ex-results} shows the experimental results, and Figure \ref{fig:training_curve} shows the corresponding test accuracy during training. As shown, the accuracy of the non-I.I.D. setting is lower than that of the I.I.D. setting, which is consistent with findings reported in prior work \cite{hsieh2019non}. 

\begin{table*}[!ht]
\centering
 \caption{Training time with FedAvg on modern CNNs (Hardware: 8 x NVIDIA Quadro RTX 5000 GPU (16GB/GPU); RAM: 512G; CPU: Intel Xeon Gold 5220R 2.20GHz).}
\label{tab:perf-results}
\resizebox{0.8\linewidth}{!}{
\begin{threeparttable}
    \begin{tabular}{lccc}
    \toprule
      \textbf{} & \textbf{ResNet-56} & \textbf{MobileNet} \\
      \midrule
      Number of workers &10&10\\
      Single-GPU standalone simulation (wall clock time) & > 4 days & > 3 days\\
      Multi-GPU distributed training (wall clock time) &11 hours&7 hours\\
       \bottomrule
    \end{tabular}
    \begin{tablenotes}[para,flushleft]
      \footnotesize
      \item *Note that the number of workers can be larger than the number of GPUs because \texttt{FedML} supports multiple processing training in a single GPU. 
    \end{tablenotes}
\end{threeparttable}}
\end{table*}
We also compared the training time of distributed computing with that of standalone simulation. The result in Table \ref{tab:perf-results} reveals that when training large CNNs, the standalone simulation is about 8 times slower than distributed computing with 10 parallel workers. Therefore, when training large DNNs, we suggest using \texttt{FedML}'s distributed computing paradigm, which is not supported by existing FL libraries such as PySyft \cite{ryffel2018generic}, LEAF \cite{caldas2018leaf}, and TTF \cite{TFF2019}. Moreover, \texttt{FedML} supports multiprocessing in a single GPU card which  enables \texttt{FedML} to run a large number of training workers by using only a few GPU cards. As an example, when training ResNet on CIFAR-10, \texttt{FedML} can run 112 workers in a server with 8 GPUs.

\vspace{-1mm}
\section{Conclusion}
\vspace{-1mm}
\texttt{FedML} is a research-oriented federated learning library and benchmark. We hope it could provide researchers and engineers with an end-to-end toolkit to facilitate developing FL algorithms and fairly comparing with existing algorithms. 
We welcome any useful feedback from the readers, and will continuously update \texttt{FedML} to support the research of the federated learning community.

\small
\bibliographystyle{nips}
\bibliography{references}

\appendix
\newpage

\section{The Taxonomy of Research Areas and a Comprehensive Publication List}

\vspace{-6mm}
\begin{table*}[htb!]
\caption{The taxonomy of research areas in federated learning and related publication statistics}
    \label{tab:taxonomy}
    \resizebox{\textwidth}{!}{
    \begin{threeparttable}
     \centering
    \begin{tabular}{p{0.17\columnwidth}p{0.75\columnwidth}cp{0.08\columnwidth}}
    \toprule
    Research Areas  & \makecell[c]{Approaches or Sub-problems (\# of Papers)} & Subtotal \\
    \midrule
    \multirow{3}{*}{\textbf{\tabincell{l}{Statistical\\ Challenges}}} &
    Distributed Optimization (56), Non-IID and Model Personalization (49), Vertical FL (8), Decentralized FL (3), Hierarchical FL (7), Neural Architecture Search (4), Transfer Learning (11), Semi-Supervised Learning (3), Meta Learning (3) & \multirow{3}{*}{144}\\
       \midrule         
      \textbf{Trustworthiness} & 
      Preserving Privacy (35), Adversarial Attack (43), Fairness (4), Incentive Mechanism (5) & 87\\
    \midrule 
    \multirow{2}{*}{\textbf{\tabincell{l}{System\\ Challenges}}} & 
    Communication-Efficiency (27), Computation Efficiency (17), Wireless Communication and Cloud Computing (71), FL System Design (19) & \multirow{2}{*}{134}\\
    \midrule 
    \multirow{2}{*}{\tabincell{l}{Models and \\Applications}} & 
    Models (22), Natural language Processing (15), Computer Vision (3), Health Care (27), Transportation (13), Other (21) & \multirow{2}{*}{101}\\
    \midrule 
   Common & Benchmark and Dataset (20), Survey (7) &27\\
    \bottomrule
    \end{tabular}
    \begin{tablenotes}
      \footnotesize
      \item From a comprehensive FL publication list: \url{https://github.com/chaoyanghe/Awesome-Federated-Learning}
\end{tablenotes}
    \end{threeparttable}
    }
\end{table*}

\vspace{-4mm}
\section{Benchmark}
\vspace{-2mm}

\subsection{Details of Supported Algorithms}
\label{appendix:alg}

\textbf{Federated Averaging (FedAvg).} FedAvg \cite{mcmahan2017communication} is a standard federated learning algorithm that is normally used as a baseline for advanced algorithm comparison. 
	We summarize the algorithm message flow in Figure \ref{fig:example_fedavg}. Each worker trains its local model for several epochs, then updates its local model to the server. The server aggregates the uploaded client models into a global model by weighted coordinate-wise averaging (the weights are determined by the number of data points on each worker locally), and then synchronizes the global model back to all workers. In our \texttt{FedML} library, based on the worker-oriented programming, we can implement this algorithm in a distributed computing manner. We suggest that users start from FedAvg to learn using \texttt{FedML}.
	
\textbf{Decentralized FL.} We use \cite{he2019central}, a central server free FL algorithm, to demonstrate how \texttt{FedML} supports decentralized topology with directed communication. As Figure \ref{fig:example_decentralized_fl} shows, such an algorithm uses a decentralized topology, and more specifically, some workers do not send messages (model) to all of their neighbors. The worker-oriented programming interface can easily meet this requirement since it allows users to define any behavior for each worker. 

\textbf{Vertical Federated Learning (VFL).} 
	VFL or feature-partitioned FL ~\cite{fmlyangqiang} is applicable to the cases where all participating parties share the same sample space but differ in the feature space. As illustrated in Figure \ref{fig:example_vfl}, VFL is the process of aggregating different features and computing the training loss and gradients in a privacy-preserving manner to build a model with data from all parties collaboratively ~\cite{Hardy2017securelr, cheng2019secureboost, liu2020backdoor, liu2020efficient}. The \texttt{FedML} library currently supports the logistic regression model with customizable local feature extractors in the vertical FL setting, and it provides NUS-WIDE \cite{Chua2009nuswide} and lending club loan \cite{lendingclubkaggle} datasets for the experiments.
	
 \textbf{Split Learning.} Split learning is computing and memory-efficient variant of FL introduced in  \cite{gupta2018distributed, vepakomma2018split} where the model is split at a layer and the parts of the model preceding and succeeding this layer are shared across the worker and server, respectively. Only the activations and gradients from a single layer are communicated in split learning, as against that the weights of the entire model are communicated in federated learning. Split learning achieves better communication-efficiency under several settings, as shown in \cite{singh2019detailed}. Applications of this model to wireless edge devices are described in  
	\cite{koda2020communication,park2019wireless}. Split learning also enables matching client-side model
    components with the best server-side model components for automating model selection as shown in work on
    ExpertMatcher \cite{sharma2019expertmatcher}.
    
\textbf{Federated Neural Architecture Search (FedNAS).} FedNAS \cite{he2020fednas} is a federated neural architecture search algorithm \cite{he2020milenas} that enables scattered clients to collaboratively search for a neural architecture. FedNAS differs from other FL algorithms in that it exchanges information beyond gradient even though it has a centralized topology similar to FedAvg. 

\subsection{Details of Datasets}
\label{appendix:dataset}
 \textbf{Federated EMNIST}: EMNIST \cite{cohen2017emnist} consists of images of digits and upper and lower case English characters, with 62 total classes. The federated version of EMNIST \cite{caldas2018leaf} partitions the digits by their author. The dataset has natural heterogeneity stemming from the writing style of each person. 

 \textbf{CIFAR-100}: Google introduced a federated version of CIFAR-100 \cite{krizhevsky2009learning} by randomly partitioning the training data among 500 clients, with each client receiving 100 examples \cite{reddi2020adaptive}. The partition method is Pachinko Allocation Method (PAM) \cite{li2006pachinko}.

 \textbf{Shakespeare}: \cite{mcmahan2017communication} first introduced this dataset to FL community. It is a dataset built from \textit{The Complete Works of William Shakespeare}. Each speaking role in each play is considered a different device. 

 \textbf{StackOverflow} \cite{TTFstackoverflow}: Google TensorFlow Federated (TFF) team maintains this federated dataset, which is derived from the Stack Overflow Data hosted by kaggle.com. We integrate this dataset into our benchmark.
 
\textbf{CIFAR-10 and CIFAR-100.} CIFAR-10 and CIFAR-100 \cite{krizhevsky2009learning} both consists of 32$\times$32 colorr images. CIFAR-10 has 10 classes, while CIFAR-100 has 100 classes. Following \cite{yurochkin2019bayesian} and \cite{wang2020federated}, we use latent Dirichlet allocation (LDA) to partition the dataset according to the number of workers involved in training in each round.

\textbf{CINIC-10.} CINIC-10 \cite{darlow2018cinic} has 4.5 times as many images as that of CIFAR-10. It is constructed from two different sources: ImageNet and CIFAR-10. It is not guaranteed that the constituent elements are drawn from the same distribution. This characteristic fits for federated learning because we can evaluate how well models cope with samples drawn from similar but not identical distributions.

\subsection{Lack of Fair Comparison: Diverse Non-I.I.D. Datasets and Models}
\begin{table*}[!htb]
\centering
\caption{various datasets and models used in latest publications from the machine learning community}
\resizebox{\linewidth}{!}{
    \begin{threeparttable}
    \begin{tabular}{cp{0.4\columnwidth}cccc}\\
    \toprule
      \textbf{Conference} & \textbf{Paper Title}  &   \textbf{dataset}  & 
      \textbf{partition method}  &
      \textbf{model} &
      \textbf{worker/device} \\
       & & & & &\textbf{number} \\
        \midrule
      \multirow{2}{*}{ICML 2019} & 
      \multirow{2}{0.4\columnwidth}{Analyzing Federated Learning through an Adversarial Lens \cite{bhagoji2019analyzing}}&
      Fashion-MNIST & natural non-IID  &3 layer CNNs & 10\\
      \cmidrule(lr){3-6}
      & & UCI Adult Census datase &   -&fully connected neural network & 10\\
       \midrule
      \multirow{4}{*}{ICML 2019} & 
      \multirow{4}{0.4\columnwidth}{Agnostic Federated Learning \cite{mohri2019agnostic}}&
      UCI Adult Census datase & -  &logistic regression & 10\\
      \cmidrule(lr){3-6}
      & & Fashion-MNIST & -  &logistic regression & 10\\
      \cmidrule(lr){3-6}
      & & Cornell movie dataset &-   &two-layer LSTM mode & 10\\
      \cmidrule(lr){3-6}
      & & Penn TreeBank (PTB) dataset & - &two-layer LSTM mode & 10\\
      \midrule
      \multirow{2}{*}{ICML 2019} & 
      \multirow{2}{0.4\columnwidth}{Bayesian Nonparametric Federated Learning of Neural Networks \cite{yurochkin2019bayesian}}&
      MNIST & Dir(0.5)  &
      1 hidden layer neural networks & 10\\
      \cmidrule(lr){3-6}
      & & CIFAR10 & Dir(0.5) &
      1 hidden layer neural networks & 10\\
      \midrule
      \multirow{6}{*}{ICML 2020} & \multirow{6}{0.4\columnwidth}{
      Adaptive Federated Optimization \cite{reddi2020adaptive} } &
      CIFAR-100 & Pachinko Allocation Method &
      ResNet-18  &  10\\
      \cmidrule(lr){3-6}
     &  & FEMNIST &natural non-IID & CNN (2xconv) & 10\\
     \cmidrule(lr){3-6}
     &  & FEMNIST& natural non-IID  & Auto Encoder & 10\\
     \cmidrule(lr){3-6}
     &  & Shakespeare & natural non-IID  & RNN & 10 \\
     \cmidrule(lr){3-6}
     &  & StackOverflow & natural non-IID  & logistic regression& 10 \\
     \cmidrule(lr){3-6}
    &  & StackOverflow & natural non-IID  &1 RNN LSTM & 10 \\
    \midrule
      \multirow{3}{*}{ICML 2020} & 
      \multirow{3}{0.4\columnwidth}{FetchSGD: Communication-Efficient Federated Learning with Sketching \cite{rothchild_fetchsgd_2020}}&
      CIFAR-10/100 & 1 class / 1 client &
      ResNet-9 &- \\
      \cmidrule(lr){3-6}
      &  & FEMNIST & natural non-IID  & ResNet-101& - \\
      \cmidrule(lr){3-6}
     &  & PersonaChat & natural non-IID  & GPT2-small &-\\
     \midrule
      \multirow{5}{*}{ICML 2020} & 
      \multirow{5}{0.4\columnwidth}{Federated Learning with Only Positive Labels \cite{yu2020federated}} &
      CIFAR-10 & 1 class / client  &
      ResNet-8/32 & -\\
      \cmidrule(lr){3-6}
       & & CIFAR-100 & 1 class / client   & ResNet-56& - \\
       \cmidrule(lr){3-6}
      & & AmazonCAT &  1 class / client  & Fully Connected Nets & -\\
      \cmidrule(lr){3-6}
      & & WikiLSHTC &  1 class / client   & - &- \\
      \cmidrule(lr){3-6}
      & & Amazon670K &   1 class / client & - & -\\
      \midrule
      \multirow{2}{*}{ICML 2020} & 
      \multirow{2}{0.4\columnwidth}{SCAFFOLD: Stochastic Controlled Averaging for Federated Learning\cite{karimireddy2019scaffold}}&
      \multirow{2}{*}{EMNIST} & \multirow{2}{*}{ 1 class / 1 client}  &\multirow{2}{*}{
      Fully connected network} & \multirow{2}{*}{-}\\
      & & & & & \\
      \midrule
      \multirow{1}{*}{ICML 2020} & 
      \multirow{1}{0.4\columnwidth}{From Local SGD to Local Fixed-Point Methods for Federated Learning\cite{malinovsky2020local}}& a9a(LIBSVM) & - & Logistic Regression &- \\
      \cmidrule(lr){3-6}
       & & a9a(LIBSVM) & - & Logistic Regression &- \\
       \midrule
      \multirow{4}{*}{ICML 2020} & 
      \multirow{4}{0.4\columnwidth}{Acceleration for Compressed Gradient Descent in Distributed and Federated Optimization\cite{li2020acceleration}}&
      a5a &  - & logistic regression & -\\
      \cmidrule(lr){3-6}
      & & mushrooms & - & logistic regression &- \\
      \cmidrule(lr){3-6}
      & & a9a & - & logistic regression & -\\
      \cmidrule(lr){3-6}
      & & w6a LIBSVM &-  & logistic regression & -\\
      \midrule
      \multirow{2}{*}{ICLR 2020} & 
      \multirow{2}{0.4\columnwidth}{Federated Learning with Matched Averaging \cite{wang2020federated}} &
      CIFAR-10 & - &
      VGG-9 & 16\\
      \cmidrule(lr){3-6}
      & & Shakespheare & sampling 66 clients  & 1-layer LSTM &66 \\
      \midrule
      \multirow{4}{*}{ICLR 2020} & 
      \multirow{4}{0.4\columnwidth}{Fair Resource Allocation in Federated Learning \cite{li2019fair}}&
      Synthetic dataset use LR & natural non-IID  &multinomial logistic regression & 10\\
\cmidrule(lr){3-6}
      & & Vehicle & natural non-IID  &SVM for binary classification & 10\\
      \cmidrule(lr){3-6}
      & & Shakespeare & natural non-IID  &RNN & 10\\
      \cmidrule(lr){3-6}
      & & Sent140 & natural non-IID  &RNN & 10\\
      \midrule
      \multirow{2}{*}{ICLR 2020} & 
      \multirow{2}{0.4\columnwidth}{On the Convergence of FedAvg on Non-IID Data\cite{li2019convergence}}&
      MNIST & natural non-IID  &logistic regression & 10\\
      \cmidrule(lr){3-6}
      & & Synthetic dataset use LR & natural non-IID  &logistic regression & 10\\
      \midrule
      \multirow{4}{*}{ICLR 2020} &
      \multirow{4}{0.4\columnwidth}{DBA: Distributed Backdoor Attacks against Federated Learning\cite{xie2019dba}}&
      Lending Club Loan Data & -  &3 FC & 10\\
      \cmidrule(lr){3-6}
      & & MNIST & -  &2 conv and 2 fc & 10\\
      \cmidrule(lr){3-6}
      & & CIFAR-10  & - &lightweight Resnet-18 & 10\\
      \cmidrule(lr){3-6}
      & & Tiny-imagenet & -  &Resnet-18 & 10\\
      \midrule
      \multirow{4}{*}{MLSys2020} & 
      \multirow{4}{0.4\columnwidth}{Federated Optimization in Heterogeneous Networks\cite{Sahu2018OnTC}} &
      MNIST & natural non-IID & multinomial logistic regression & 10\\
      \cmidrule(lr){3-6}
      & & FEMNIST & natural non-IID  &multinomial logistic regression & 10\\
      \cmidrule(lr){3-6}
      & & Shakespeare & natural non-IID  &RNN & 10\\
      \cmidrule(lr){3-6}
      & & Sent140 & natural non-IID  &RNN & 10\\
      \bottomrule
    \end{tabular}
    \begin{tablenotes}[para,flushleft]
      \footnotesize
      \item *Note: we will update this list once new publications are released.
    \end{tablenotes}
    \end{threeparttable}}
\label{tab:stat-datasets}
\end{table*}

\section{IoT Devices}
\label{iot_devices}
Currently, we support two IoT devices: Raspberry PI 4 (Edge CPU Computing) and NVIDIA Jetson Nano (Edge GPU Computing).

\subsection{Raspberry Pi 4 (Edge CPU Computing - ARMv7l)}


Raspberry Pi 4 Desktop kit is supplied with:

\begin{itemize}
  \item Raspberry Pi 4 Model B (2GB, 4GB or 8GB version)
  \item Raspberry Pi Keyboard and Mouse
  \item 2 × micro HDMI to Standard HDMI (A/M) 1m Cables
  \item Raspberry Pi 15.3W USB-C Power Supply
  \item 16GB NOOBS with Raspberry Pi OS microSD card
\end{itemize}
For more details, please check this link: \ \url{https://www.raspberrypi.org/products/raspberry-pi-4-desktop-kit}.

\subsection{NVIDIA Jetson Nano (Edge GPU Computing)}
NVIDIA® Jetson Nano™ Developer Kit is a small, powerful computer that lets you run multiple neural networks in parallel for applications like image classification, object detection, segmentation, and speech processing. All in an easy-to-use platform that runs in as little as 5 watts. 

For more details, please check this link: \ \url{https://developer.nvidia.com/embedded/jetson-nano-developer-kit}.


\end{document}